\documentclass{article}

\usepackage[main, final]{neurips_2026}
\usepackage{amsmath}
\usepackage{mathtools}
\usepackage{tcolorbox}
\tcbuselibrary{skins,breakable} 
\usepackage[table]{xcolor}   
\usepackage{booktabs}        
\usepackage{multirow}
\usepackage{xcolor}
\usepackage{colortbl}
\usepackage{svg}
\definecolor{background}{HTML}{EDE7DE}
\usepackage{array}
\usepackage{subcaption}
\usepackage{siunitx}
\usepackage[table]{xcolor}
\usepackage{wrapfig}
\usepackage{float}
\usepackage[framemethod=default]{mdframed}   

\newcommand{\eat}[1]{\ignorespaces}

\newcommand{\aautoref}[1]{\hyperref[#1]{Appendix~\ref*{#1}}}

\definecolor{winrow}{HTML}{F5F1E6}    
\definecolor{groupline}{gray}{0.75}

\makeatletter
\AtBeginDocument{%
  \renewcommand\paragraph{%
    \@startsection{paragraph}{4}{\z@}%
                  {0.5ex \@plus 0.3ex \@minus 0pt}
                  {-1em}
                  {\normalsize\bf}%
  }%
}
\makeatother

\newcounter{takeaway}
\newtcolorbox{takeawaybox}{%
  enhanced, breakable, parbox=false,
  colback=background!75!white,
  colframe=background!50!black,
  left=4pt, right=4pt, top=2pt, bottom=2pt,
  leftrule=3pt, rightrule=0pt, toprule=0pt, bottomrule=0pt,
  arc=3pt,
  before skip=5pt, after skip=10pt,
}

\newcommand{\newtakeaway}[1]{%
  \refstepcounter{takeaway}%
  \begin{takeawaybox}%
    \textbf{Takeaway \thetakeaway:}\enspace #1%
  \end{takeawaybox}%
}

\newcounter{finding}

\newtcolorbox{findingheadingbox}{%
  enhanced,
  colback=background!35!white,
  colframe=background!55!black,
  boxrule=0pt,
  leftrule=2.5pt,
  rightrule=0pt,
  toprule=0pt,
  bottomrule=0pt,
  arc=1pt,
  left=6pt,
  right=6pt,
  top=3pt,
  bottom=3pt,
  before skip=10pt,
  after skip=5pt,
}

\newcommand{\newhypothesis}[2]{\begin{tcolorbox}[colback=background!75!white, colframe=background!50!black,
  left=1pt, right=1pt, top=1pt, bottom=1pt,
  leftrule=3pt, rightrule=0pt, toprule=0pt, bottomrule=0pt]
{{\textbf{Hypothesis: \emph{#1}}\par #2}}
\end{tcolorbox}
}

\newcolumntype{A}{>{\columncolor{gray!10}}c}  
\newcolumntype{G}{>{\columncolor{sizecell}}c}

\definecolor{winrow}{RGB}{235, 244, 252}   
\definecolor{sizecell}{RGB}{226, 232, 240} 
\definecolor{groupline}{gray}{0.7}

\usepackage{xspace}

\newcommand{\Standard}{\textsc{Standard}\xspace}
\newcommand{\SPS}{\textsc{SPS}\xspace}
\newcommand{\DelayedState}{\textsc{Delayed State}\xspace}
\newcommand{\Twomem}{\textsc{2x Memory}\xspace}   
\newcommand{\Inverse}{\textsc{Reverse SPS}\xspace}   

\usepackage[utf8]{inputenc} 
\usepackage[T1]{fontenc}    
\usepackage{hyperref}       
\usepackage{url}            
\usepackage{booktabs}       
\usepackage{amsfonts}       
\usepackage{nicefrac}       
\usepackage{microtype}      
\usepackage{xcolor}         

\title{The State-Prediction Separation Hypothesis}

\author{%
  Giovanni Monea$^\dagger$ \\
  \And
  Nathan Godey$^\dagger$ \\
  \And
  Kianté Brantley$^\diamond$ \\
  \And
  Yoav Artzi$^\dagger$ \\ \\
}

\begin{document}

\maketitle

\vspace{-30pt} 
\noindent\makebox[\linewidth][c]{%
    $^\dagger$Cornell University \quad $^\diamond$Harvard University%
}

\noindent\makebox[\linewidth][c]{%
  \texttt{giovanni@cs.cornell.edu, \{ng554, yoavartzi\}@cornell.edu}%
}

\vspace{-5pt} 
\noindent\makebox[\linewidth][c]{%
  \texttt{kdbrantley@g.harvard.edu}%
}
\vspace{5pt}

\begin{abstract}

Transformers use the same forward computation stream to both predict the next token and store useful state for future token predictions.
We formulate the \emph{state-prediction separation hypothesis}: disentangling the two roles yields better language modeling performance.
We design a Transformer variant that uses two computation streams to separate the two functions, and conduct pretraining experiments across various scales. Our experiments show that state-prediction separation consistently offers better data and compute efficiencies, improving validation loss and outperforming standard Transformers by 2--3 percentage points on average on downstream tasks. We also conduct extensive empirical analysis that rules out potential confounders and demonstrates the fundamental difference in the gradients our design entails.

\end{abstract}

\begin{figure}[h!]
    \centering
    \begin{subfigure}{\textwidth}
        \centering
        \includegraphics[width=\textwidth, trim={0 0.1cm 0 0.1cm}, clip]{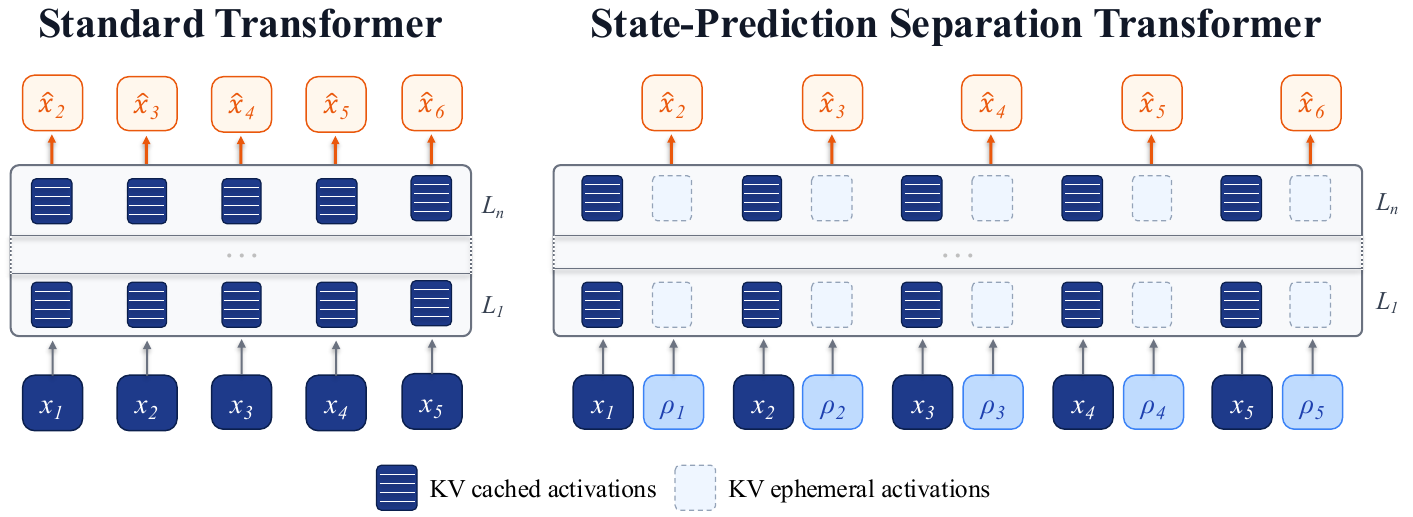}
        \label{fig:transformers_comparison}
    \end{subfigure}

    \vspace{-1em}

    \begin{subfigure}{\textwidth}
        \centering
        \includegraphics[width=\textwidth, trim={0 0.5cm 0 0.1cm}, clip]{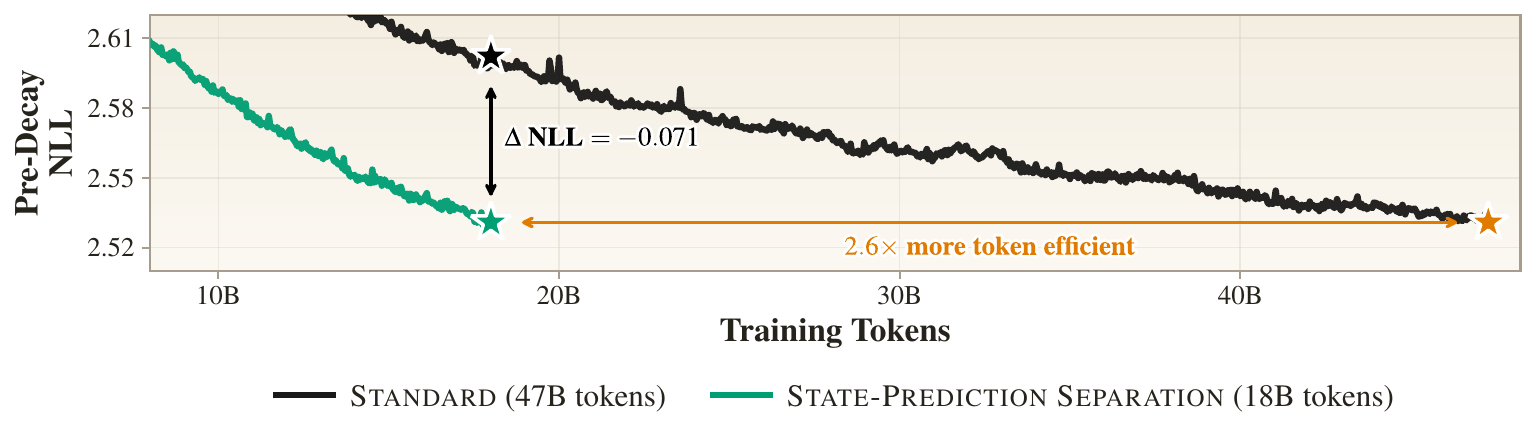}
        \label{fig:abstract_pre_decay_nll}
    \end{subfigure}
    \vspace{-2em}
    \caption{\textbf{Standard versus State-Prediction Separation Transformer.} \textit{Top:} The standard Transformer uses the same hidden states for both memory and prediction. Our variant separates these roles: input token $x_i$ time steps form a persistent state, while prediction token steps $\rho_i$ produce next-token predictions. \textit{Bottom:} At 1.6B parameters, State-Prediction Separation matches the validation loss of a standard Transformer trained on $47$B tokens while using $2.6\times$ fewer tokens  (pre-decay). At an $18$B-token pre-decay budget, it already achieves $\Delta\mathrm{NLL}=-0.071$ versus standard.}
    \label{fig:transformers_overview}
    \vspace{-10em}
\end{figure}

\newpage


\section{Introduction}\label{sec:intro}

Attention-based architectures, including the Transformer~\citep{NIPS2017_3f5ee243} and earlier recurrent designs~\citep{DBLP:journals/corr/BahdanauCB14}, have dual use for the activations computed at each time step: they are used to predict the output of that time step (i.e., token in language models) and are attended to by subsequent steps. 
The first role is focused on prediction; the second on capturing state information to be reused later on.
Generally, these two functions are entangled in the same representation and computation \emph{stream} (i.e., forward path). 
In this paper, we propose and study the following hypothesis in large language models (LLMs):
\newhypothesis{State-Prediction Separation (simplified)}{The next-token prediction computation and state representation compete when forced through the same computation. Routing them through separate streams yields better language modeling.}

Technically, we separate the state and prediction functionalities by inserting an additional computation time step before predicting the next token (\autoref{fig:transformers_overview}). Time steps then appear in pairs: first, the token previously generated is processed, but no new token is emitted. An additional time step follows, which emits the next token to generate. The key-value (KV) entries from the first of the two steps are added to the KV cache, while the entries of the latter of the two are discarded.\footnote{In practice, as we describe in \autoref{sec:sps-transformer}, we retain prediction KV activations within a sliding window.}
This design distinguishes between two streams: a state stream and a prediction stream.

We conduct extensive experiments pretraining a set of LLMs at common research scales, from 53M to 1.678B parameters. 
The main result is that SPS significantly improves pretraining performance over standard Transformers in both token-equivalent and compute-equivalent settings.
\autoref{fig:transformers_overview} illustrates one of the key results: non-separating baselines cannot match the training loss of an SPS Transformer even with double the number of training tokens. 
We also show that state-prediction separation outperforms several variants controlling for SPS's compute and memory overheads, proving that separation is the key component in the improvements we report.

Our code is available at \url{https://github.com/lil-lab/sps}.

\section{Prediction and State Preparation}

We consider a standard autoregressive Transformer with vocabulary $V$, depth $L$, and parameters $\theta$. 
\autoref{app:transformer} details the full architecture.
The input is a sequence $x = (x_1,\ldots,x_T)$. 
At sequence position $i$ with input token $x_i$, the model computes a per-layer hidden state $h_i^{(l)} \in \mathbb{R}^d$, where $l=1,\dots,L$, $h_i^{(0)}$ is the token embedding, and $h_i^{(L)}$ the final representation. Each position $i$ also contributes entries to the key-value (KV) cache.
%
Through causal attention, all the past keys and values (i.e., from positions $k < i$) contribute to the final representation $h_i^{(L)}$, which is then used to compute the next-token distribution.
Each hidden state $h_i^{(l)}$ plays two roles: it is part of the computation of the immediate prediction for $x_{i+1}$, and it produces KV entries read by every later position. 

This double role is reflected in the optimization gradients, as they are computed through backpropagation. 
The language modeling training loss is $\mathcal{L} = \frac{1}{T-1}\sum_{i=1}^{T-1} \ell_i$, with the per-position cross-entropy loss $\ell_i = -\log p(x_{i+1}\mid x_{\le i})$. 
The parameters $\theta$ are used repeatedly in each position $i=1,\dots,T$ in the Transformer. 
We isolate the gradients for each position $i$ by denoting $\nabla_{\theta_i}\mathcal{L}$, and can similarly denote $\nabla_{\theta_i}\ell_j$ to denote the gradients for position $i$ from the loss at position $j$.\footnote{This follows how the forward pass creates a rolled-out computation graph with the parameters $\theta$ used repeatedly.} 
The gradients of the loss $\mathcal{L}$ by $\theta$ are the linear sum of all per-position gradients:
\begin{equation}
    \nabla_\theta\mathcal{L} \;=\; \sum_{i=1}^{T-1} \nabla_{\theta_i}\mathcal{L} \;=\; \frac{1}{T-1}\sum_{i=1}^{T-1}\sum_{j=1}^{T-1}\nabla_{\theta_i}\ell_j \;=\; \frac{1}{T-1}\sum_{i=1}^{T-1}\sum_{j=i}^{T-1}\nabla_{\theta_i}\ell_j \;\;.\label{eq:causal-grad}
\end{equation}
The last term follows pruning every non-causal $\ell_j$, because of the causality in attention, each step's parameters affect only the current and future losses. 
Separating the step's own loss ($j=i$) from the losses back-propagated only through the KV cache ($j>i$) decomposes the gradient by source:
\begin{equation}\label{eq:aggregate-grad}
    \nabla_{\theta}\mathcal{L}
    \;=\; \sum_{i=1}^{T-1}\;[\;
    \underbrace{\vphantom{\sum_{i=j+1}^{T-1}}\tfrac{1}{T-1}\nabla_{\theta_i}\ell_i}_{\text{Prediction}}
    \;+\;
    \underbrace{\tfrac{1}{T-1}\sum_{j=i+1}^{T-1} \nabla_{\theta_i}\ell_j }_{\text{State}}\;]\;\;.
\end{equation}

Time step $i$ contributes gradients for the prediction of $x_{i+1}$\,---\,the \emph{prediction} task\,---\,and for the preparation of keys and values that help all later positions $j>i$ make better predictions\,---\,the \emph{state representation} task. Both components flow (i.e., back-propagate) through the same hidden state $h_i^{(l)}$, which is therefore optimized to conflate the two roles in a single set of activations.

\section{The State-Prediction Separation Transformer}\label{sec:sps-transformer}

We separate the two roles by augmenting the standard Transformer with an additional learned token, \texttt{<predict>}, inserted after every input token. 
Given an input sequence $x=(x_1,\ldots,x_T)$, we form an augmented sequence by interleaving dummy tokens $\rho_i$, all set to a new \texttt{<predict>} token:
\begin{equation}\label{eq:interleaved}
    x \rightarrow (x_1,\rho_1,\; x_2,\rho_2,\; \ldots,\; x_T,\rho_T)\;\;.
\end{equation}
The two tokens $x_i$ and $\rho_i$ at index $i$ share the same position encoding. The model now maintains two interleaved streams of representations: an \emph{input stream} $\{x_i\}_{i=1}^T$ that we use to carry the state forward, and a \emph{prediction stream} $\{\rho_i\}_{i=1}^T$ that we use to emit next-token predictions. 
Beyond a sliding window of size $w$, only key-value elements from the input stream positions are available in the KV cache, to be attended by later positions. 
The sliding window allows to attend to the specific token-choice representations for a short horizon (i.e., for local coherence). 
The prediction $x_{i+1}$ is done at the position of $\rho_i$, so the loss is applied only at $\rho_i$ positions:
\begin{equation}\label{eq:sps-loss}
    \mathcal{L}
    =
    \frac{1}{T-1}\sum_{i=1}^{T-1}
    -\log p\bigl(x_{i+1} \mid x_1,\rho_1,\ldots,x_i,\rho_i\bigr)\;\;.
\end{equation}

In a standard Transformer, the two streams are tied together: $\rho_i$ does not exist, and the same representations $h_i^{(l)}$ at each position must simultaneously pack the information to emit the prediction for $x_{i+1}$ and produce the keys and values read by every later position. The two gradient components of \autoref{eq:aggregate-grad} (the prediction term $\nabla_{\theta_i}\ell_i$ and the state-preparation term $\sum_{j>i}\nabla_{\theta_i}\ell_j$) are routed through one and the same $h_i^{(\ell)}$, with no architectural separation. \autoref{fig:transformers_overview} illustrates the State-Prediction Separation Transformer (\SPS), and compares it to the standard architecture.

We can now make the simplified hypothesis from \autoref{sec:intro} precise in this two-stream notation:


\newhypothesis{State-Prediction Separation}{The prediction gradients  $\nabla_{\theta_{\rho_i}}\ell_i$ and the state-preparation gradients $\sum_{j>i}\nabla_{\theta_{x_i}}\ell_j$ are both used to optimize the computation of the same representations $h_i$, thereby competing with each other in the standard Transformer. Separating the hidden representations to $h_{\rho_i}$ and $h_{x_i}$ and routing the gradients appropriately separates the two functions, and yields lower next-token loss at matched parameter count.}

\begin{figure}[t]
    \centering
    \includegraphics[width=\textwidth]{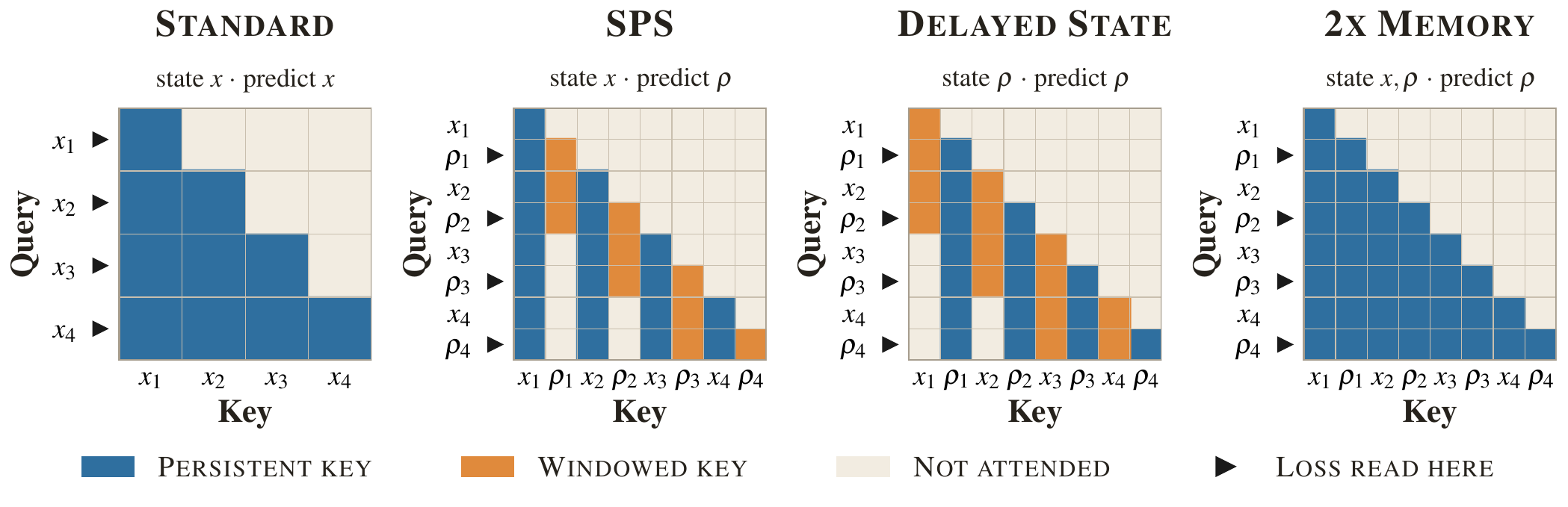}
    \caption{\textbf{Attention mask and prediction slot of each method.} Rows are queries and columns are keys, drawn with $T{=}4$ tokens and window $w{=}1$. Persistent keys are never evicted, while windowed keys are visible only while they remain within $w$. A marker on the left of a row indicates that the next-token loss is read from that slot.}
    \label{fig:attention_masks_main}
\end{figure}

At training time, we realize this separation through an attention mask. In \SPS, input entries are persistent, while \texttt{<predict>} entries are evicted once they leave a sliding window of size $w$. A query $q$ at step $i$ (i.e., either $x_i$ or $\rho_i$) attends to all causal input entries and only recent \texttt{<predict>} entries. \autoref{fig:attention_masks_main} illustrates this mask. 
The only difference between input and prediction positions is that prediction positions attend to their corresponding input position: 
\begin{equation}\label{eq:sps-mask}
    \mathcal{A}_{\mathrm{SPS}}(i,q)
    =
    \underbrace{\{x_k : k \leq i\}}_{\text{All causal inputs}}
    \;\cup\;
    \underbrace{
    \begin{cases}
        \{\rho_k : i-w \leq k < i\} & \text{if } q=x_i\\
        \{\rho_k : i-w \leq k \leq i\} & \text{if } q=\rho_i
    \end{cases}}_{\text{Recent \texttt{<predict>} entries}}\;\;.
\end{equation}
The persistent KV cache of \SPS contains only input entries; \texttt{<predict>} entries are read by at most $w$ later queries before being discarded. This routes the two gradient components of \autoref{eq:aggregate-grad} to different streams. Input representations $h_{x_i}^{(l)}$ are visible to every later query, so they accumulate the full state-preparation gradient $\sum_{j>i}\nabla_{h_{x_i}^{(l)}}\ell_j$. \texttt{<predict>} representations $h_{\rho_i}^{(l)}$ are visible only within a window. Their gradient is dominated by the immediate prediction term $\nabla_{\theta_{\rho_i}}\ell_i$, with a contribution limited to at most $w-1$ following state-preparation losses.
\autoref{fig:transformers_overview} contrasts a standard Transformer with our \SPS Transformer.

\paragraph{Training Efficiency}
Our method increases compute in order to separate the state and prediction streams, which makes training more expensive due to the doubled context length. We efficiently simulate the sliding window through attention masking, and apply the same mechanism to prevent attention from crossing document boundaries. Due to the additional computations, a single training step takes up to 2$\times$ as long as with \Standard (see \autoref{tab:main-compact} and \autoref{app:flops}). Nevertheless, as we demonstrate below, \SPS still requires less overall time to reach a given target model quality because it learns from fewer tokens.

\paragraph{Inference Efficiency}
The additional cost is negligible at inference. Forwarding one or a few tokens simultaneously incurs essentially the same latency. This is a well-known property that motivates speculative decoding~\citep{chen2023acceleratinglargelanguagemodel, leviathan2023fastinferencetransformersspeculative}. Concretely, \SPS's persistent KV cache contains only input tokens, matching the size of a standard Transformer cache, with a bounded $w$-slot ring buffer holding the most recent \texttt{<predict>} entries. Each generated token triggers a single decode step that forwards the pair $(x_i, \rho_i)$ jointly and reads next-token logits from the \texttt{<predict>} hidden state.


\section{Experimental Setup}\label{sec:experimental-setup}

\paragraph{Baselines}
We compare \SPS\ to a standard Transformer (\Standard) and to two ablations. The first, \Twomem, retains both input and \texttt{<predict>} entries in the KV cache throughout the sequence. The model gains computational capacity from its doubled context length, but its persistent memory footprint also doubles, and \texttt{<predict>} entries still serve both prediction and state-preparation. 

The second, \DelayedState, inserts a \texttt{<predict>} token after every input, giving the model an extra computation step before each prediction, but commits the persistent state at the \texttt{<predict>} slot, one step \emph{after} the input. Compared to \SPS, this variant delays state preparation as well, and performs both prediction and state preparation together at the \texttt{<predict>} slot. A query $q$ at step $i$ attends to the $w$ most recent input entries,
using the same fixed-size ephemeral window as \SPS\ but populated by input
entries rather than \texttt{<predict>} entries, and to all causal \texttt{<predict>} entries.

\DelayedState\ keeps the persistent KV cache size roughly equivalent to \Standard's. However, unlike \SPS, no separation between roles is enforced. The \texttt{<predict>} stream carries both prediction and state. \DelayedState\ therefore differs from \SPS\ only in \emph{whether} the two streams are separated. \autoref{fig:attention_masks_main} contrasts the attention pattern and prediction slot of every variant.

\begin{wraptable}{r}{0.55\textwidth}
      \vspace{-1.0em}
      \centering
      \small
      \setlength{\tabcolsep}{4pt}
      \begin{tabular}{l c c c c c}
          \toprule
          & \textbf{XS} & \textbf{S} & \textbf{M} & \textbf{L} & \textbf{XL} \\
          \midrule
          Layers ($L$)            & 8    & 12   & 24    & 36   & 48    \\
          Hidden size ($d$)       & 512  & 768  & 1024  & 1280 & 1600  \\
          Heads ($H$)             & 8    & 12   & 16    & 20   & 25    \\
          FFN size                & 1536 & 2304 & 3072  & 3840 & 4800  \\
          Parameters              & 53M  & 131M & 379M  & 831M & 1.678B \\
          \bottomrule
      \end{tabular}
      \caption{\textbf{Model configurations across the five scales.}}
      \label{tab:architecture-scales}
      \vspace{-1.0em}
  \end{wraptable}
All variants share the same backbone, a pre-normalized Transformer blocks with RMSNorm~\citep{NEURIPS2019_1e8a1942}, SwiGLU feed-forward networks~\citep{shazeer2020gluvariantsimprovetransformer} of intermediate size $3d$, rotary positional embeddings~\citep{su2021roformer}, no biases on linear layers, a weight-tied unembedding, and a context length of $4{,}096$ tokens. We evaluate five scales, summarized in \autoref{tab:architecture-scales}. For S, M, L, and XL, we follow the GPT-2~\citep{radford2019language} recipe, and we add XS as a smaller scale. \SPS, \Twomem, and \DelayedState use the same backbone, parameter count, and hyperparameters as \Standard\ at every scale, differing only in attention pattern. Unless otherwise specified, all sliding-window variants use $w{=}64$ at every scale.

\paragraph{Data}
We pretrain on FineWeb-Edu~\citep{penedo2024the}, a high-quality educational subset of FineWeb, and tokenize with the GPT-2 tokenizer. Sequences are packed across document boundaries, with an end-of-sequence token (\texttt{<eos>}) inserted between consecutive documents to delimit them. Attention is masked so that queries within a document cannot attend to keys from any other document, and we exclude \texttt{<eos>} positions from the next-token loss so that the model is never trained to predict the start of an unrelated document. Each model is trained for $20$B tokens by default. This budget meets or exceeds the Chinchilla compute-optimal ratio of ${\approx}20$ tokens per parameter~\citep{hoffmann2022an} at every scale, except for XL (which is under-trained due to its higher cost of training).\footnote{Chinchilla-optimal $\approx$ $1.1$B, $2.6$B, $7.6$B, $16.6$B, and $33.6$B tokens for XS, S, M, L, and XL.} To obtain a fair GPU-hours comparison, we additionally train \Standard for $40$B tokens (except for XL, that is trained for 47B tokens, until it matches \SPS validation loss). All runs see the same data in the same order.

\paragraph{Training}
We base our training code on nanoGPT~\citep{Karpathy2022}, including its standard hyperparameters.\footnote{AdamW with $\beta_1{=}0.9$, $\beta_2{=}0.95$, weight decay $0.1$, gradient clipping at $1.0$, and peak learning rate $6\!\times\!10^{-4}$.} The global batch size is $96$ sequences of length $4{,}096$, i.e.,\ $\approx\!400{,}000$ tokens per gradient update. All models are trained in bfloat16 mixed precision. We use a learning-rate schedule~\citep{hagele2024scaling} consisting of a brief linear warmup, a constant phase at the peak learning rate, and a linear decay covering the final $10\%$ of training tokens. For faster training, we adapt the open-source Triton implementation of FlashAttention~\citep{dao2022flashattention} to support the sliding-window and \texttt{<predict>}-token attention patterns of \SPS, \DelayedState, and \Twomem, so that all variants train at comparable throughput; we found alternatives such as FlexAttention~\citep{dong2024flexattentionprogrammingmodel} to be either memory-inefficient or substantially slower in our setting. All XS/S/M/L runs use $2$ NVIDIA H100 $80$\,GB GPUs, while all XL runs use $2$ NVIDIA B200 GPUs, with data-parallel distributed training. All runs in the main results use a single seed (i.e., the data ordering and weight-initialization seed are matched across methods).

\paragraph{Evaluation}
We report three families of metrics. \emph{(a) Validation loss.} Held-out NLL on FineWeb-Edu, our pretraining distribution. \emph{(b) Generalization.} Corpus NLL averaged over four out-of-distribution corpora (WikiText~\citep{merity2017pointer}, C4~\citep{10.5555/3455716.3455856}, Pile-Books3~\citep{pile}, GovReport~\citep{huang-etal-2021-efficient}), and zero-shot accuracy averaged over five standard benchmarks (ARC-Easy~\citep{clark2018thinksolvedquestionanswering}, HellaSwag~\citep{zellers-etal-2019-hellaswag}, PIQA~\citep{DBLP:conf/aaai/BiskZLGC20}, SciQ~\citep{welbl-etal-2017-crowdsourcing}, LAMBADA~\citep{paperno-etal-2016-lambada}), evaluated as standard practice via the LM Evaluation Harness~\citep{eval-harness}. \emph{(c) Inference efficiency.} End-to-end throughput (tokens/s) and peak GPU memory measured on a single NVIDIA H100 for a batch of $16$ sequences with a prefill of $1024$ tokens followed by $3072$ decode steps, reported as ratios relative to \Standard at the same scale. Every method runs through the same generation loop, KV-cache layout, and Triton attention path, with each method using the fused kernel matched to its own attention pattern. We additionally report forward FLOPs for a full $4{,}096$-token prefill relative to \Standard, computed analytically in \autoref{app:flops}, as it does not depend on our specific implementations.

\section{Results}

\begin{figure}[t!]
    \centering
    \includegraphics[width=\textwidth]{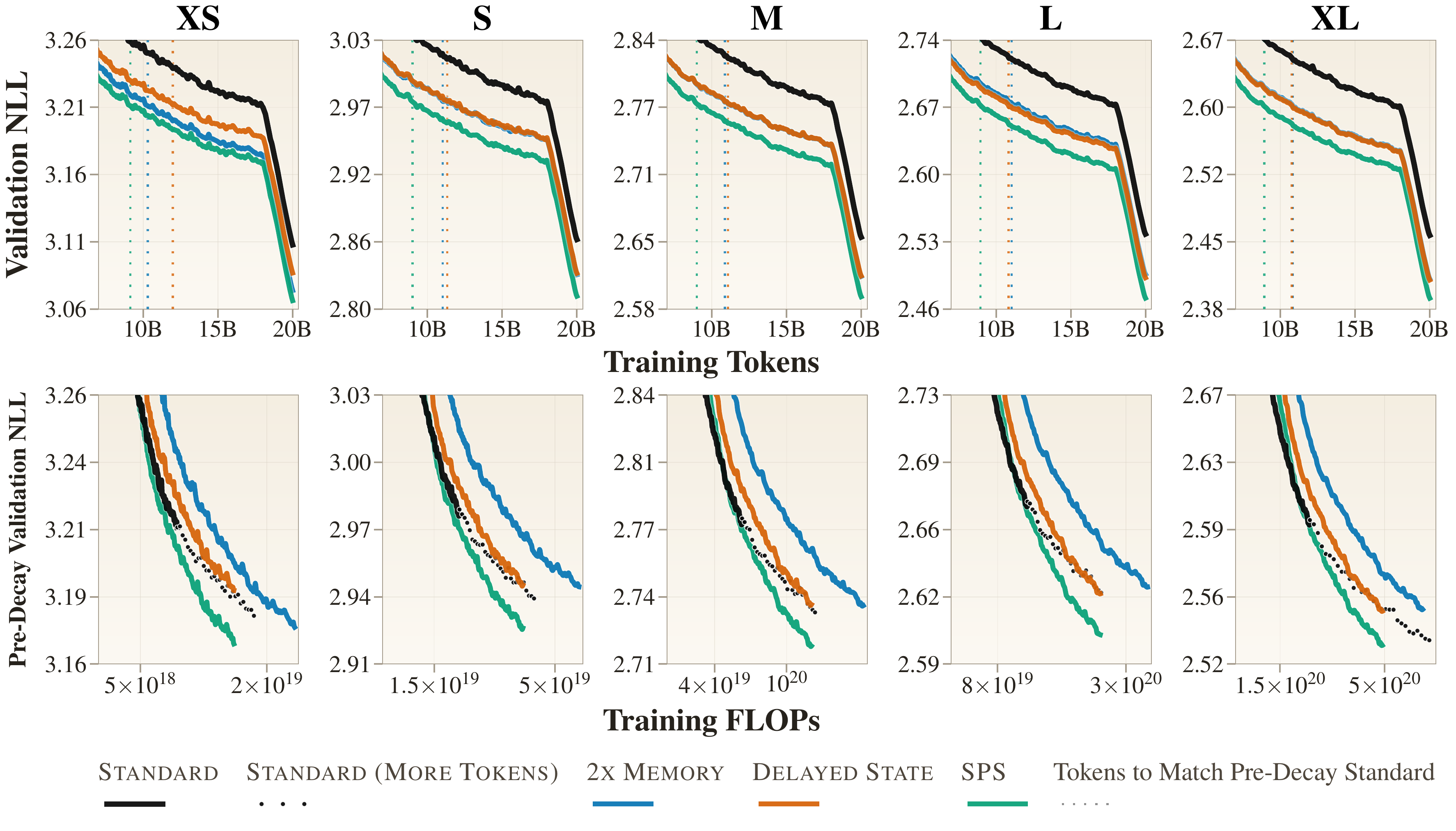}
    \caption{\textbf{\SPS\ trains faster and reaches lower loss at every scale.} FineWeb-Edu validation NLL vs.\ tokens seen (top) and FLOPs (bottom). The top row includes LR cool down, hence the drop at the end. The bottom row is pre-decay only.}
    \label{fig:history_combined}
\end{figure}

\begin{table}[ht]
    \centering
    \vspace*{2em}
    \setlength{\tabcolsep}{6pt}
    \renewcommand{\arraystretch}{1.2}
    \resizebox{\textwidth}{!}{%
    \begin{tabular}{@{}c l
        S[table-format=1.3]
        S[table-format=1.3] S[table-format=2.1]
        S[table-format=1.2] S[table-format=1.2] S[table-format=1.2]@{}}
        \toprule
        & & {\textsc{Validation Loss}}
        & \multicolumn{2}{c}{\textsc{Generalization}}
        & \multicolumn{3}{c}{\textsc{Efficiency}} \\
        \cmidrule(lr){3-3} \cmidrule(lr){4-5} \cmidrule(lr){6-8}
        Size & Method
        & {FW-Edu ($\downarrow$)}
        & {Corpus NLL ($\downarrow$)}
        & {Task Acc. (\%, $\uparrow$)}
        & {Throughput ($\times$, $\uparrow$)}
        & {Peak Mem. ($\times$, $\downarrow$)}
        & {FLOPs ($\times$, $\downarrow$)} \\
        \midrule
        \multirow{4}{*}{\textbf{XS}}
         & \Standard & 3.107 & 4.335 & 44.9 & 1.00 & 1.00 & 1.00 \\
         & \Twomem & {3.073\,\scriptsize {($-0.034$)}} & {4.311\,\scriptsize {($-0.024$)}} & {47.3\,\scriptsize {($+2.4$)}} & {0.93\,\scriptsize {($-0.07$)}} & {1.81\,\scriptsize {($+81\%$)}} & 2.48 \\
         & \DelayedState & {3.086\,\scriptsize {($-0.021$)}} & {4.312\,\scriptsize {($-0.023$)}} & {46.9\,\scriptsize {($+2.0$)}} & {0.94\,\scriptsize {($-0.06$)}} & {1.01\,\scriptsize {($+1\%$)}} & 1.72 \\
        \rowcolor{winrow}
         & \SPS & {{\bfseries 3.065}\,\scriptsize {($-0.042$)}} & {{\bfseries 4.243}\,\scriptsize {($-0.092$)}} & {{\bfseries 47.3}\,\scriptsize {($+2.5$)}} & {0.94\,\scriptsize {($-0.06$)}} & {1.01\,\scriptsize {($+1\%$)}} & 1.72 \\
        \addlinespace[3pt]
        \arrayrulecolor{groupline}\cmidrule[0.3pt](l{2pt}r{2pt}){1-8}\arrayrulecolor{black}
        \addlinespace[3pt]
        \multirow{4}{*}{\textbf{S}}
         & \Standard & 2.858 & 4.018 & 49.5 & 1.00 & 1.00 & 1.00 \\
         & \Twomem & {2.829\,\scriptsize {($-0.030$)}} & {3.981\,\scriptsize {($-0.036$)}} & {51.1\,\scriptsize {($+1.5$)}} & {0.93\,\scriptsize {($-0.07$)}} & {1.81\,\scriptsize {($+81\%$)}} & 2.55 \\
         & \DelayedState & {2.829\,\scriptsize {($-0.029$)}} & {3.992\,\scriptsize {($-0.026$)}} & {51.8\,\scriptsize {($+2.2$)}} & {0.94\,\scriptsize {($-0.06$)}} & {1.01\,\scriptsize {($+1\%$)}} & 1.83 \\
        \rowcolor{winrow}
         & \SPS & {{\bfseries 2.810}\,\scriptsize {($-0.048$)}} & {{\bfseries 3.913}\,\scriptsize {($-0.105$)}} & {{\bfseries 51.8}\,\scriptsize {($+2.3$)}} & {0.94\,\scriptsize {($-0.06$)}} & {1.01\,\scriptsize {($+1\%$)}} & 1.83 \\
        \addlinespace[3pt]
        \arrayrulecolor{groupline}\cmidrule[0.3pt](l{2pt}r{2pt}){1-8}\arrayrulecolor{black}
        \addlinespace[3pt]
        \multirow{4}{*}{\textbf{M}}
         & \Standard & 2.648 & 3.732 & 55.8 & 1.00 & 1.00 & 1.00 \\
         & \Twomem & {2.610\,\scriptsize {($-0.038$)}} & {3.687\,\scriptsize {($-0.045$)}} & {57.3\,\scriptsize {($+1.4$)}} & {0.93\,\scriptsize {($-0.07$)}} & {1.81\,\scriptsize {($+81\%$)}} & 2.61 \\
         & \DelayedState & {2.611\,\scriptsize {($-0.037$)}} & {3.701\,\scriptsize {($-0.031$)}} & {57.5\,\scriptsize {($+1.7$)}} & {0.95\,\scriptsize {($-0.05$)}} & {1.01\,\scriptsize {($+1\%$)}} & 1.92 \\
        \rowcolor{winrow}
         & \SPS & {{\bfseries 2.591}\,\scriptsize {($-0.058$)}} & {{\bfseries 3.627}\,\scriptsize {($-0.106$)}} & {{\bfseries 58.7}\,\scriptsize {($+2.9$)}} & {0.95\,\scriptsize {($-0.05$)}} & {1.01\,\scriptsize {($+1\%$)}} & 1.92 \\
        \addlinespace[3pt]
        \arrayrulecolor{groupline}\cmidrule[0.3pt](l{2pt}r{2pt}){1-8}\arrayrulecolor{black}
        \addlinespace[3pt]
        \multirow{4}{*}{\textbf{L}}
         & \Standard & 2.537 & 3.598 & 60.1 & 1.00 & 1.00 & 1.00 \\
         & \Twomem & {2.495\,\scriptsize {($-0.042$)}} & {3.542\,\scriptsize {($-0.055$)}} & {62.1\,\scriptsize {($+1.9$)}} & {0.83\,\scriptsize {($-0.17$)}} & {1.78\,\scriptsize {($+78\%$)}} & 2.58 \\
         & \DelayedState & {2.491\,\scriptsize {($-0.045$)}} & {3.532\,\scriptsize {($-0.066$)}} & {61.3\,\scriptsize {($+1.2$)}} & {0.90\,\scriptsize {($-0.10$)}} & {1.01\,\scriptsize {($+1\%$)}} & 1.96 \\
        \rowcolor{winrow}
         & \SPS & {{\bfseries 2.470}\,\scriptsize {($-0.067$)}} & {{\bfseries 3.484}\,\scriptsize {($-0.113$)}} & {{\bfseries 62.6}\,\scriptsize {($+2.5$)}} & {0.90\,\scriptsize {($-0.10$)}} & {1.01\,\scriptsize {($+1\%$)}} & 1.96 \\
        \addlinespace[3pt]
        \arrayrulecolor{groupline}\cmidrule[0.3pt](l{2pt}r{2pt}){1-8}\arrayrulecolor{black}
        \addlinespace[3pt]
        \multirow{4}{*}{\textbf{XL}}
         & \Standard & 2.458 & 3.487 & 63.2 & 1.00 & 1.00 & 1.00 \\
         & \Twomem & {2.411\,\scriptsize {($-0.047$)}} & {3.423\,\scriptsize {($-0.064$)}} & {64.3\,\scriptsize {($+1.0$)}} & {0.64\,\scriptsize {($-0.36$)}} & {1.75\,\scriptsize {($+75\%$)}} & 2.52 \\
         & \DelayedState & {2.410\,\scriptsize {($-0.048$)}} & {3.433\,\scriptsize {($-0.055$)}} & {64.8\,\scriptsize {($+1.6$)}} & {0.94\,\scriptsize {($-0.06$)}} & {1.01\,\scriptsize {($+1\%$)}} & 1.97 \\
        \rowcolor{winrow}
         & \SPS & {{\bfseries 2.390}\,\scriptsize {($-0.068$)}} & {{\bfseries 3.338}\,\scriptsize {($-0.149$)}} & {{\bfseries 66.3}\,\scriptsize {($+3.1$)}} & {0.94\,\scriptsize {($-0.06$)}} & {1.01\,\scriptsize {($+1\%$)}} & 1.97 \\
        \bottomrule
    \end{tabular}%
    }
    \vspace*{2em}
    \caption{\textbf{\SPS outperforms all baselines on quality while remaining comparable to \Standard in memory and throughput.} Main results across XS--XL. \textbf{Bold} marks the best per column within each size; \SPS rows are shaded. Task accuracy averages 5 zero-shot benchmarks; Corpus NLL averages 4 corpora. Throughput is end-to-end tokens/sec for a combined prefill 1k + decode 3k workload at batch size 16 relative to \Standard on H100; Peak Memory is the ratio of peak GPU memory used during decode. FLOPs is forward FLOPs for a full 4096-token prefill relative to \Standard. \autoref{tab:main-full} details tasks and corpora performance.}
    \label{tab:main-compact}
\end{table}

\paragraph{Performance and Efficiency}\label{sect:sps_vs_standard}

\begin{figure}[t!]
    \centering
    \begin{minipage}[t]{0.48\textwidth}
        \centering
        \includegraphics[width=\linewidth]{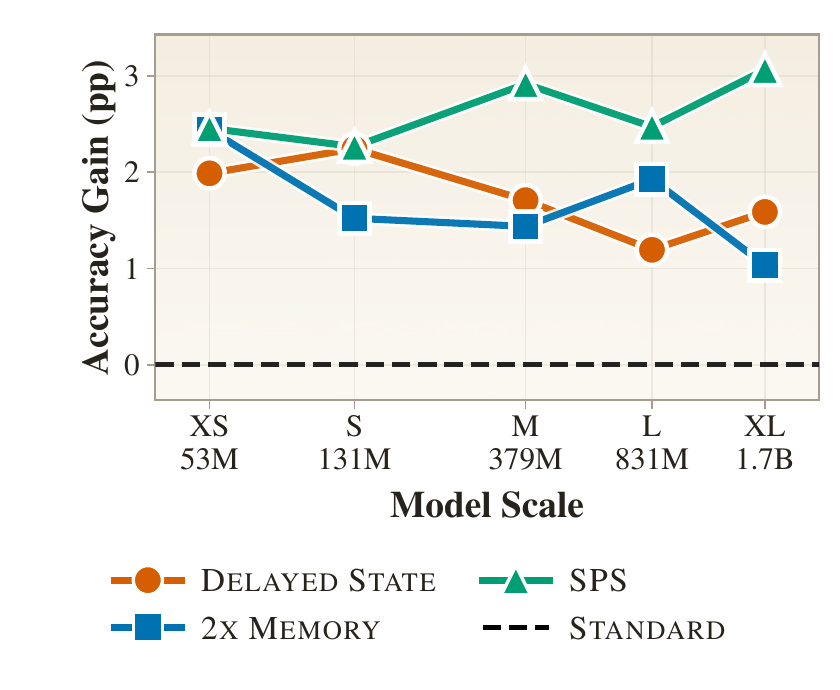}
        \caption{\textbf{\SPS\ gives consistent downstream accuracy gains, with the largest observed gain at the largest scale.} \SPS\ improves average accuracy at every scale, with gains of roughly $2$--$3$ percentage points and the largest gain observed at XL.}
        \label{fig:accuracy_gains}
    \end{minipage}
    \hfill
    \begin{minipage}[t]{0.48\textwidth}
        \centering
        \includegraphics[width=\linewidth]{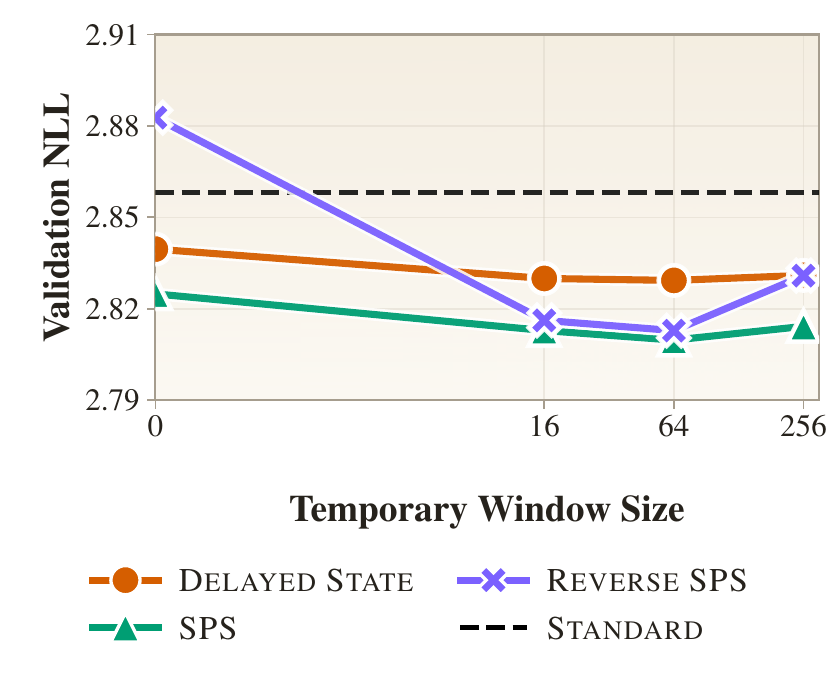}
        \caption{\textbf{\SPS\ works best at small but non-zero $w$, and outperforms \Inverse\ at every window.} Final FineWeb-Edu validation NLL vs.\ ephemeral window size for \SPS, \DelayedState, and Reverse \SPS\ at S, after $20$B training tokens.}
        \label{fig:window_ablation}
    \end{minipage}
\end{figure}

\autoref{tab:main-compact} summarizes performance and efficiency findings. 
\SPS\ attains the lowest validation NLL on FineWeb-Edu at every scale, with the gap over \Standard\ widening from $-0.042$ at XS to $-0.068$ at XL. \autoref{fig:history_combined} shows that at matched training tokens \SPS\ reaches lower validation NLL than \Standard\ throughout training (top), and that at matched FLOPs (and GPU hours, \autoref{app:flops}) \SPS\ eventually overtakes \Standard\ at every scale (bottom). Even doubling \Standard's pre-decay training budget from $18$B to $36$B tokens does not close the gap. The same holds when the budget is spent on a larger \Standard\ model instead: beyond $2.1\times10^{20}$ FLOPs, \SPS's compute-optimal frontier leads (\autoref{app:compute-optimal}). \SPS\ thus reaches \Standard's quality on roughly half the training data, with the data-efficiency ratio widening as scale grows. This is an important property as high-quality human-generated text approaches projected exhaustion~\citep{villalobos2024run}. The improvement carries over to held-out generalization. Corpus NLL on four out-of-distribution corpora drops by $0.09$--$0.11$ across scales, and zero-shot accuracy on five standard benchmarks improves by $2.3$--$3.1\%$. \autoref{fig:accuracy_gains} shows this trend directly. 
Crucially, this quality gain comes at minimal increase in inference cost. \SPS's persistent KV cache is the same size as \Standard's (peak memory ratio $1.01$), and end-to-end throughput is within $6$--$10\%$ of \Standard\ at all scales.  While each result above is from a single training run, we verify with a $3$-seed sweep at S, $10$B that \SPS's gap over \Standard, \DelayedState, and \Twomem\ is significant at $p<0.005$ (one-sided Welch's $t$-test; \autoref{app:seeds}).

\newtakeaway{\SPS\ outperforms \Standard\ in validation loss, generalization, and learning speed at every scale, matching \Standard's persistent memory footprint and inference throughput while reaching \Standard's quality on roughly half the training data.}

\paragraph{State and Prediction Role Separation}\label{sect:pred_vs_mem}

\Twomem\ keeps every \texttt{<predict>} entry persistent at the same per-step compute as \SPS, which discards \texttt{<predict>} entries beyond the window. \SPS\ is consistently better in validation NLL across XS--XL, even though it has half the persistent KV cache. This shows that the gain is not capacity-based. Keeping \texttt{<predict>} entries persistent forces each one to serve both as a prediction site \emph{and} as a state carrier for later queries, re-coupling the two streams that \SPS\ separates.

A second hypothesis is that \SPS\ benefits only from giving the model an additional Transformer step before the persistent state is committed. \DelayedState\ tests this directly. It has the same per-step compute as \SPS\ and the same persistent-cache size as \Standard\ and \SPS, but commits the persistent state at the \texttt{<predict>} slot, one step \emph{after} the input, and so does not separate the two roles. \DelayedState\ does improve over \Standard, confirming that the extra computation step carries some benefit, but at every scale \SPS\ remains consistently better, by $0.019$--$0.021$ in validation NLL and by $0.06$--$0.07$ in corpus NLL (\autoref{tab:main-compact}).


\newtakeaway{The specific structure of state-prediction separation in \SPS matters more than extra computations or even simply doubling the memory.}

\paragraph{Which Stream Should Persist, and at What Window?}
\label{sect:window_ablation}

We additionally run a \Inverse variant that swaps the two roles. \texttt{<predict>} entries form the persistent state, and input entries are windowed and used to emit the next-token prediction. This is the mirror image of \SPS, which predicts from \texttt{<predict>} and stores state from input (\autoref{fig:attention_masks}).
This isolates whether the specific role assignment matters. We jointly ablate the \texttt{<predict>}-window size $w$ and the choice of persistent stream at the S scale, where a sweep is cheap to run, and reuse the resulting $w$ across all scales in the main experiments (\autoref{fig:window_ablation}). Both \SPS\ and \DelayedState\ show nearly constant performance across all $w \in \{0, 16, 64, 256\}$, with $w{=}64$ as the empirical best by a small margin; we therefore fix $w{=}64$ at every scale. \Inverse matches \SPS\ at moderate $w$ but degrades sharply at small $w$, where windowing the input stream cuts off recent input visibility. \SPS's ordering is the more robust default: persisting inputs tolerate a wider range of $w$ before quality drops.

\newtakeaway{\SPS\ is robust to window size, while \Inverse\ collapses at small $w$. Persisting the input stream is the more forgiving design.}

\begin{figure}[t!]
    \centering
    \begin{subfigure}[b]{\textwidth}
        \centering
        \includegraphics[width=\textwidth]{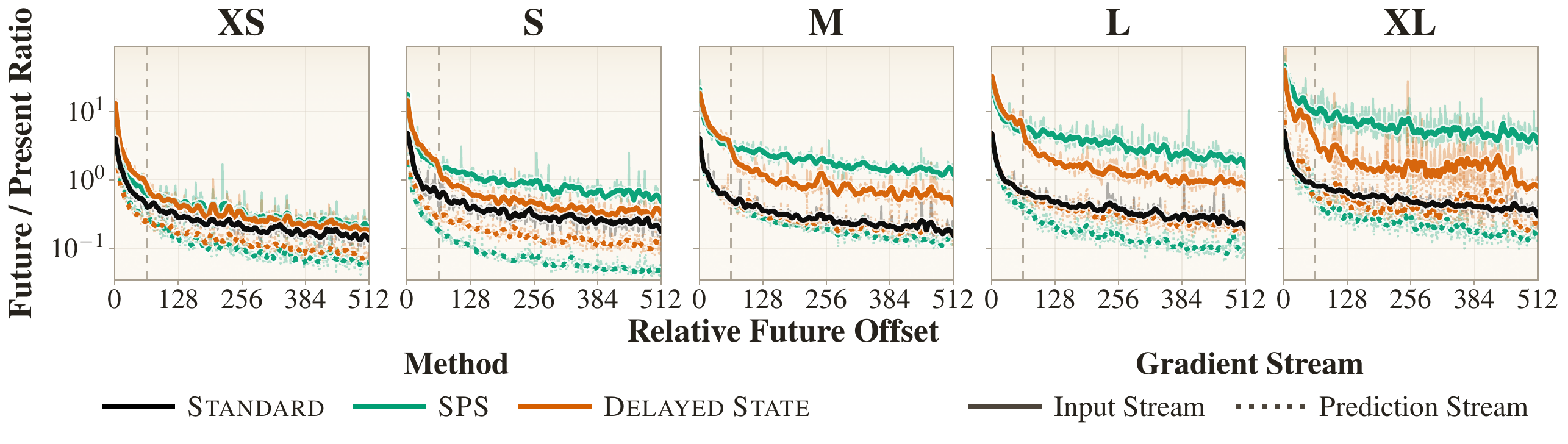}
        \caption{\textbf{Per-offset gradient ratio $r(p,\,k)$ for \Standard, \SPS, and \DelayedState .} For \SPS and \DelayedState, solid curves are the input stream ($p{=}x_i$) and dotted curves the prediction stream ($p{=}\rho_i$). Bold curves are Savitzky--Golay--smothed~\citep{doi:10.1021/ac60214a047} trends. The faint curves underneath are the raw per-offset means. \SPS's input stream consistently sustains more future-loss gradient; \DelayedState's prediction stream stays low, and its input stream collapses past the ephemeral window $k{=}64$.}
        \label{fig:analysis_gradient}
    \end{subfigure}
    \vskip\baselineskip
    \begin{subfigure}[b]{\textwidth}
        \centering
        \includegraphics[width=\textwidth]{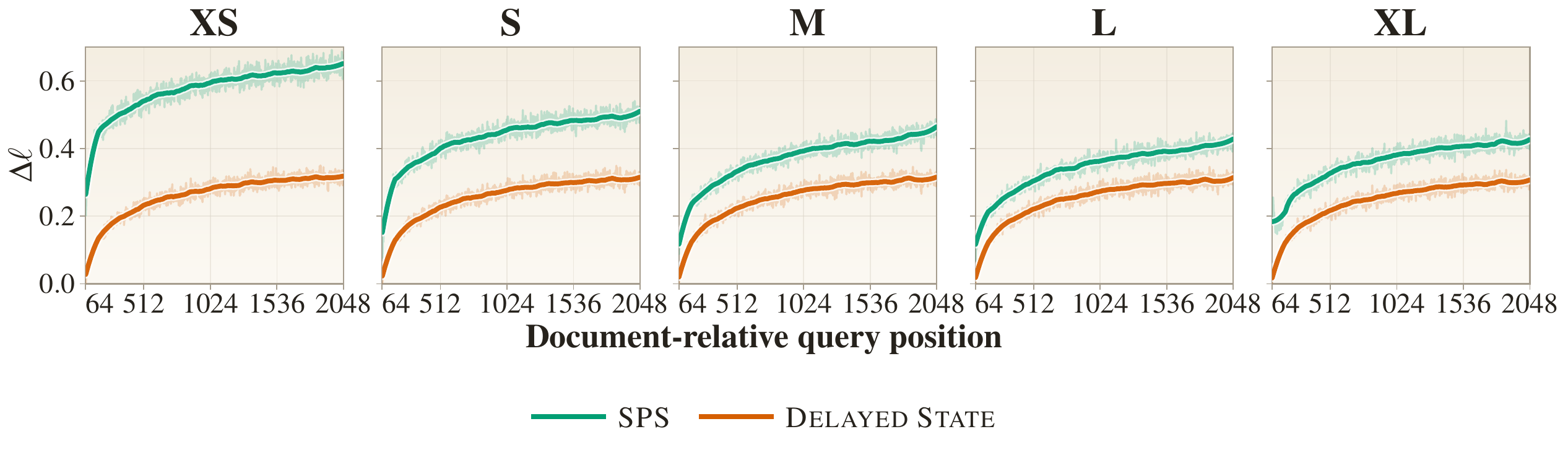}
        \caption{\textbf{Per-position NLL degradation when each method's persistent state is restricted to a window of size $w{=}64$, plotted against the document-relative query position $t$ at each scale.} Bold curves are Savitzky--Golay--smoothed~\citep{doi:10.1021/ac60214a047} trends. The faint curves underneath are the raw per query position means. \SPS's curve sits uniformly above \DelayedState's, with the late-position gap ranging from $\sim\!2.0\times$ at XS to $\sim\!1.3\times$ at XL. \SPS's persistent keys carry more future-relevant information and evicting them is more hurtful.}
        \label{fig:analysis_delta}
    \end{subfigure}
    \caption{\textbf{Analysis of \SPS.} (a) Where future-loss gradient lands during training. (b) How much the persistent state is actually used at inference.}
    \label{fig:sps_analysis}
\end{figure}

\paragraph{Analysis: Why Does Separation Help?}
Following \autoref{eq:aggregate-grad}, we probe how each architecture allocates gradients between the \emph{prediction} role (the immediate loss $\ell_i$) and the \emph{state} representation role (future losses $\ell_j$, $j > i$) at training time, and what this implies at inference. 
Recall that we denote as $\nabla_i \ell_j$ the gradients from the loss term $\ell_j$ due to the use of the parameters in position $i$. 

In \SPS\ and \DelayedState\ each step $i$ occupies two positions in the interleaved sequence: an input slot $x_i$ where the input token is read, and a predict slot $\rho_i$ where the cross-entropy $\ell_i$ is computed (the only position where a prediction happens).
Both positions use the parameters $\theta$. Therefore, we can separate the gradients to $\nabla_{x_i}\ell_j$ and $\nabla_{\rho_i}\ell_j$.
Because $x_i$'s hidden states feed into $\rho_i$, both slots receive non-zero gradient from $\ell_i$ even though the loss is realized only at $\rho_i$. 
This is in contrast to \Standard, where each step occupies a single position with gradients $\nabla_i \ell _j$.



For every source position $p$ at step $i$ (so $p \in \{x_i, \rho_i\}$ in \SPS\ and \DelayedState, $p = x_i$ in \Standard), we isolate $\nabla_p \ell_{i+k}$, the gradients from the step-$k$-ahead loss, and compute the ratio\footnote{We exclude the language-model head from $\theta_p$, since it carries only the present-loss prediction role we are trying to isolate.}
\begin{equation}
    r(p,\,k) \;=\; \frac{\bigl\lVert \nabla_{\theta_p}\, \ell_{i+k} \bigr\rVert_2}{\bigl\lVert \nabla_{\theta_p}\, \ell_i \bigr\rVert_2}\;\;,
\end{equation}
the magnitude of position-$p$ gradients coming from the loss $k$ steps ahead relative to the current time step gradients. We average $r$ over $8{,}000$ examples ($1{,}000$ documents, $8$ source positions each) for $k \le 512$, separately on input positions ($p{=}x_i$) and prediction positions ($p{=}\rho_i$) for \SPS and \DelayedState. \autoref{fig:analysis_gradient} shows a clean dichotomy. \SPS's input stream carries more future-loss gradient than \Standard at every offset, and its prediction stream carries strictly less, meaning the two roles are routed to different tokens. \DelayedState reduces this separation. Its prediction stream stays uniformly low, and its input stream falls below \SPS past the \texttt{<predict>} window $k{=}64$, beyond which gradient can only flow indirectly. A single stream does not allow the model to effectively learn to predict and to represent state.

Carrying future-loss gradient is necessary but not sufficient, the persistent state must also be \emph{important} at inference. We test this by restricting each variant's persistent state to a sliding window of size $\omega$ (distinct from the prediction window $w$) and measuring the resulting NLL degradation. For a trained model $M \in \{\SPS,\,\DelayedState\}$, we define $M_\omega$ as $M$ used with a forced sliding persistent-cache of size $\omega$, $\ell_i(M)$ as the loss of the vanilla model $M$ at position $i$, and $\ell_i(M_\omega)$ is the loss of the altered $M_\omega$ at the same position $i$. We measure 
\[
    \Delta\ell_i(M_\omega) \;=\; \ell_i(M_\omega) - \ell_i(M),
\]
for document positions $i \in \{1,\dots,2048\}$. We set a small $\omega=64$ (as opposed to the vanilla $4{,}096$) and average each $\Delta\ell_i$ across $8{,}000$ documents. A larger $\Delta\ell_i$ means more of $M$'s long-range prediction depends on persistent keys outside the $\omega$-window. \autoref{fig:analysis_delta} shows that ablating \SPS's out-of-window persistent state hurts NLL $1.4$--$2.2\times$ more than ablating \DelayedState's across scales, although with the full persistent cache \SPS performs better than \DelayedState. Therefore, the future-loss gradient \SPS routes onto the input stream actually translates into a persistent state the model relies on at inference.

\newtakeaway{\SPS better separates present-loss and future-loss gradients and produces a persistent state more important at inference compared to \DelayedState.}

\section{Related Work}

\paragraph{Tension Between Present and Future Predictions}
Each hidden state in a Transformer is asked to do two jobs at once: encode the next-token prediction at its own position, and prepare the persistent state that later positions will read from. \citet{wu2024do} study this tension precisely by contrasting two hypotheses: \textit{breadcrumbs} (the keys and values useful for the current prediction also serve future ones) and \textit{pre-caching} (some computation in early positions is targeted at later predictions and would be wasted on the current one). They find pre-caching in pretrained Pythia, increasing with scale, consistent with mechanistic evidence that earlier-position representations already encode upcoming-token information~\citep{elhage2021mathematical,Pal_2023}. Like \SPS, YOCO~\citep{sun2024you} builds memory from only half of its computation, but splits by layer rather than by token: only the first half of the layers produces the KV cache, which the second half reuses. The early two-stream attention of XLNet~\citep{NEURIPS2019_dc6a7e65} also distinguishes prediction from \emph{content}, but in service of permutation language modeling rather than to relieve pre-caching under standard left-to-right training.  \SPS\ is a direct architectural response to this tension. Rather than asking one stream to serve both jobs, it inserts a dedicated \texttt{<predict>} slot at every position to carry the next-token prediction, freeing the input stream to specialize as persistent state. If the two-jobs view is right, separating the roles should help. Our experiments confirm this at every scale.

\paragraph{Adding Compute on the Input Side}
One approach to relieve this tension is adding computation at input positions. \citet{goyal2024think} append \texttt{<pause>} tokens to the prompt so the model gets extra forward passes before answering, motivated by the fact that Transformer expressivity is bounded by context length~\citep{merrill2024the}. \citet{pfau2024lets} use the same insertion as filler tokens during training. These methods share \SPS's mechanism of adding extra tokens, but use it to add capacity rather than to separate the two roles. Our \DelayedState\ and \Twomem\ baselines isolate this distinction. Both retain the inserted-token mechanism and the extra compute, but do not enable the separation, and both underperform \SPS\ at every scale. 

\paragraph{Enriching the Future-Prediction Signal}
A complementary line of work intervenes on the prediction target. \citet{pmlr-v235-bachmann24a} show that teacher-forced next-token prediction can silently fail on planning tasks where one step is hard and the rest are easy, motivating training signals that reach beyond the immediate next token. Multi-token-prediction methods~\citep{NEURIPS2018_c4127b91,monea2023pass,gloeckle2024betterfasterlarge,deepseekai2024deepseekv3technicalreport,ahn2025efficientjointpredictionmultiple,gerontopoulos2026multitoken} (and \citet{qi2020prophetnet}, who extend the two streams of \citet{NEURIPS2019_dc6a7e65} to one per future token) and belief-state objectives~\citep{hu2025the,teoh2026nextlatent} address this by adding auxiliary losses at non-current positions to strengthen the future-prediction signal itself. \SPS\ pursues a different goal. We do not enrich what is predicted, but separate where prediction and state-preparation take place. By routing the next-token loss onto a dedicated \texttt{<predict>} slot, \SPS\ relieves the present-future tension structurally, while remaining compatible with these approaches.

\section{Discussion}\label{sec:discussion}

We introduce the \SPS hypothesis that posits that the two tasks each hidden state must perform, predicting the next token and preparing state for later predictions, interfere when forced through one representation, and that separating them structurally should help. 
We study the hypothesis with the \SPS Transformer, which realizes this separation via two interleaved streams, using non-persistent states for prediction. The experiments are decisive: at every scale from XS to XL, \SPS\ lowers FineWeb-Edu validation NLL, improves held-out-corpus NLL, and raises zero-shot accuracy, at the same persistent KV-cache footprint as the standard Transformer and within a few percent of its throughput. 
Our experiments show that separation is key to the observed improvement. Our gradient-flow and restricted-state analyses confirm the mechanism: \SPS\ routes future-loss gradient onto the input stream and produces a persistent state significantly more impactful for future states than alternatives. 
\SPS\ shows dramatic data efficiency gains, which increase monotonically across XS--XL, suggesting it would only grow with more compute. This matters in a regime where high-quality data is finite and approaching projected exhaustion~\citep{villalobos2024run}. Learning more from each token directly extends the runway for pretraining.

Compute constrains the scope of our evidence in two ways. First, we pretrain on a single corpus (FineWeb-Edu); the consistent gains on out-of-distribution corpora and zero-shot benchmarks suggest the trends transfer beyond it, but we could not test alternative pretraining mixtures. Second, our largest scale is $1.678$B parameters; the \SPS-vs-\Standard\ NLL gap monotonically widens across XS--XL, suggesting the trend should continue past 1.6B, but this requires further verification. 

Our argument that mixing prediction and state-preparation in one hidden state is suboptimal rests on controlled ablations and gradient/state analyses; a formal characterization of \emph{when} and \emph{how much} this conflation hurts, as a function of capacity, depth, or data, would tighten the case and is left for future work. \SPS's prediction stream adds a forward-pass slot per input position, at most doubling per-step training compute over the standard Transformer (\autoref{app:flops}); whether the same separation can be obtained at lower overhead, via shallower or narrower computation on the prediction stream, or a sparser persistent state, is an open and practically valuable question. The two streams currently also share all parameters. We leave for future work whether further separating them (e.g., via distinct attention/FFN parameters per stream) could yield further gains now that the roles are decoupled.

\section*{Acknowledgments}

This research was partially supported by a gift to the LinkedIn–Cornell Bowers Strategic Partnership; AI-MI and NSF Award 2433348; the National Science Foundation NSF under award OAC-2311521; and NASA under award No. 20-OSTFL20-0053. NG is supported by an Empire AI Postdoctoral Fellowship. 
We gratefully acknowledge use of the research computing resources of the Empire AI Consortium, Inc., with support from the State of New York, the Simons Foundation, and the Secunda Family Foundation~\citep{10.1145/3708035.3736070}.
Any opinions, findings and conclusions or recommendations expressed in this material are those of the author(s) and do not necessarily reflect the views of the National Science Foundation or of NASA.
We thank the members of the Cornell LIL Lab for helpful discussions. 
KB acknowledges this work has been made possible in part by a gift from the Chan Zuckerberg Initiative Foundation to establish the Kempner Institute for the Study of Natural and Artificial Intelligence.

\bibliography{references}
\bibliographystyle{plainnat}


\appendix

\section{Full Transformer Notation}\label{app:transformer}

Let $V$ denote a finite vocabulary. An autoregressive Transformer defines a distribution $p \colon V^{\leq T} \to \Delta(V)$ mapping a sequence of tokens to a distribution over the next token.

The model consists of $L$ Transformer layers, with $H$ attention heads each, with dimensions $d_h = d/H$. For any given sequence position $i$, each token $x_i$ is mapped to an embedding $h_i^{(0)} = E_{x_i} \in \mathbb{R}^d$, where $E \in \mathbb{R}^{|V| \times d}$ is a learned embedding matrix (so $E_{x_i}$ denotes its $x_i$-th row).

The embeddings are processed by $L$ blocks combining causal multi-head self-attention ($\mathrm{MHA}$) with a position-wise feed-forward network ($\mathrm{FFN}$) and normalization
layers ($\mathrm{Norm}$), producing the per-layer hidden states $h_i^{(l)} \in \mathbb{R}^d$. For $l = 1, \ldots, L$:%
\begin{align*}
    \tilde{h}_i^{(l)} &= h_i^{(l-1)} + \mathrm{MHA}^{(l)}\!\bigl(
        \mathrm{Norm}^{(l)}_{\mathrm{MHA}}(h^{(l-1)})\bigr)_i,\\
    h_i^{(l)} &= \tilde{h}_i^{(l)} + \mathrm{FFN}^{(l)}\!\bigl(
        \mathrm{Norm}^{(l)}_{\mathrm{FFN}}(\tilde{h}_i^{(l)})\bigr).
\end{align*}
Let $\bar{h}_i^{(l)} \coloneqq \mathrm{Norm}^{(l)}_{\mathrm{MHA}}(h_i^{(l-1)})$ denote the normalized input to layer $l$. Each head $\eta \in \{1, \ldots, H\}$ is parameterized by projection matrices $W_Q^{(l,\eta)}, W_K^{(l,\eta)}, W_V^{(l,\eta)} \in \mathbb{R}^{d_h \times d}$ and computes
\begin{equation}\label{eq:qkv}
    q_i^{(l,\eta)} = W_Q^{(l,\eta)} \bar{h}_i^{(l)}, \qquad
    k_i^{(l,\eta)} = W_K^{(l,\eta)} \bar{h}_i^{(l)}, \qquad
    v_i^{(l,\eta)} = W_V^{(l,\eta)} \bar{h}_i^{(l)}.
\end{equation}
A rotary positional transform $\mathcal{R}_i$ is applied to queries and keys, and the per-head output is obtained by causally masked attention, then concatenated across heads and mixed by an output projection
$W_O^{(l)} \in \mathbb{R}^{d \times d}$:
\begin{align}
    o_i^{(l,\eta)} &= \sum_{j \leq i}
    \mathrm{softmax}_{j}\!\left(
        \frac{(\mathcal{R}_i q_i^{(l,\eta)})^{\!\top} (\mathcal{R}_j k_j^{(l,\eta)})}{\sqrt{d_h}}
    \right) v_j^{(l,\eta)}, \label{eq:head-out}\\
    \mathrm{MHA}^{(l)}(\bar{h}^{(l)})_i &=
    W_O^{(l)} \bigl[\, o_i^{(l,1)};\, \ldots;\, o_i^{(l,H)} \,\bigr]. \label{eq:mha}
\end{align}

The next-token distribution is obtained by applying a final RMSNorm and a weight-tied unembedding to the final representation $h_i^{(L)}$:
\begin{equation}\label{eq:unembed}
    p(\cdot \mid x_{\leq i}) = \mathrm{softmax}\!\bigl(
        E\, \mathrm{RMSNorm}_{\text{f}}(h_i^{(L)})
    \bigr).
\end{equation}

\section{Full Main Results}\label{app:main-full}

\autoref{tab:main-full} expands \autoref{tab:main-compact} with the per-corpus NLLs (WikiText, C4, Pile-Books3, GovReport) and per-benchmark zero-shot accuracies (ARC-Easy, HellaSwag, PIQA, SciQ, LAMBADA) that are averaged into Corpus NLL and Task Accuracy in the main text, along with the prefill-throughput ratio.

\begin{table}[ht]
    \centering
    \vspace*{2em}
    \footnotesize
    \setlength{\tabcolsep}{6pt}
    \renewcommand{\arraystretch}{1.2}
    \begin{tabular}{@{}c l
        S[table-format=1.3] S[table-format=1.3] S[table-format=1.3] S[table-format=1.3]
        S[table-format=2.1] S[table-format=2.1] S[table-format=2.1] S[table-format=2.1] S[table-format=2.1]@{}}
        \toprule
        & & \multicolumn{4}{c}{NLL Generalization ($\downarrow$)}
        & \multicolumn{5}{c}{Task Generalization (\%, $\uparrow$)} \\
        \cmidrule(lr){3-6} \cmidrule(lr){7-11}
        Size & Method
        & {WT} & {C4} & {Books3} & {GR}
        & {ARC-E} & {HS} & {PIQA} & {SciQ} & {LAMB} \\
        \midrule
        \multirow{4}{*}{\textbf{S}}
         & \Standard & 3.466 & 4.340 & 4.908 & 3.357 & 50.3 & 34.8 & 64.0 & 71.5 & 27.0 \\
         & \Twomem & {\underline{3.417}} & {\underline{4.302}} & {\underline{4.896}} & {\underline{3.310}} & 49.5 & 36.2 & {\underline{65.0}} & 73.4 & {\underline{31.3}} \\
         & \DelayedState & 3.423 & 4.304 & 4.928 & 3.312 & {\bfseries 51.1} & {\underline{36.5}} & 64.3 & {\bfseries 76.1} & 30.9 \\
        \rowcolor{winrow}
         & \SPS & {\bfseries 3.368} & {\bfseries 4.274} & {\bfseries 4.735} & {\bfseries 3.273} & {\underline{50.3}} & {\bfseries 37.4} & {\bfseries 65.4} & {\underline{74.1}} & {\bfseries 31.8} \\
        \addlinespace[3pt]
        \arrayrulecolor{groupline}\cmidrule[0.3pt](l{2pt}r{2pt}){1-11}\arrayrulecolor{black}
        \addlinespace[3pt]
        \multirow{4}{*}{\textbf{M}}
         & \Standard & 3.182 & 4.058 & 4.603 & 3.087 & 56.6 & 42.5 & 67.7 & 77.3 & 35.0 \\
         & \Twomem & {\underline{3.141}} & 4.026 & {\underline{4.552}} & {\underline{3.030}} & 55.9 & {\underline{44.9}} & 68.4 & 79.7 & {\underline{37.4}} \\
         & \DelayedState & 3.150 & {\underline{4.006}} & 4.618 & 3.032 & {\underline{57.0}} & 44.7 & {\underline{68.7}} & {\underline{79.9}} & 37.4 \\
        \rowcolor{winrow}
         & \SPS & {\bfseries 3.101} & {\bfseries 3.974} & {\bfseries 4.443} & {\bfseries 2.988} & {\bfseries 59.5} & {\bfseries 45.8} & {\bfseries 69.0} & {\bfseries 80.4} & {\bfseries 39.0} \\
        \addlinespace[3pt]
        \arrayrulecolor{groupline}\cmidrule[0.3pt](l{2pt}r{2pt}){1-11}\arrayrulecolor{black}
        \addlinespace[3pt]
        \multirow{4}{*}{\textbf{L}}
         & \Standard & 3.063 & 3.913 & 4.468 & 2.946 & 61.9 & 47.5 & 69.3 & 81.7 & 40.4 \\
         & \Twomem & 3.006 & 3.856 & 4.421 & 2.886 & {\bfseries 62.9} & 50.6 & 70.5 & {\bfseries 84.6} & {\underline{41.8}} \\
         & \DelayedState & {\underline{2.996}} & {\underline{3.853}} & {\underline{4.396}} & {\underline{2.881}} & 61.0 & {\underline{50.8}} & {\bfseries 71.4} & 81.9 & 41.6 \\
        \rowcolor{winrow}
         & \SPS & {\bfseries 2.953} & {\bfseries 3.830} & {\bfseries 4.313} & {\bfseries 2.841} & {\underline{62.8}} & {\bfseries 52.2} & {\underline{70.7}} & {\underline{83.1}} & {\bfseries 44.3} \\
        \addlinespace[3pt]
        \arrayrulecolor{groupline}\cmidrule[0.3pt](l{2pt}r{2pt}){1-11}\arrayrulecolor{black}
        \addlinespace[3pt]
        \multirow{4}{*}{\textbf{XL}}
         & \Standard & 2.954 & 3.812 & 4.336 & 2.846 & 64.5 & 52.6 & 71.6 & 84.2 & 43.2 \\
         & \Twomem & 2.895 & {\underline{3.749}} & {\underline{4.265}} & 2.783 & 64.0 & 55.1 & 71.6 & 84.1 & {\underline{46.6}} \\
         & \DelayedState & {\underline{2.893}} & 3.752 & 4.303 & {\underline{2.783}} & {\underline{64.7}} & {\underline{55.4}} & {\bfseries 72.0} & {\underline{85.9}} & 46.1 \\
        \rowcolor{winrow}
         & \SPS & {\bfseries 2.831} & {\bfseries 3.718} & {\bfseries 4.061} & {\bfseries 2.740} & {\bfseries 66.3} & {\bfseries 56.3} & {\underline{71.9}} & {\bfseries 87.5} & {\bfseries 49.5} \\
        \bottomrule
    \end{tabular}
    \vspace*{2em}
    \caption{\textbf{Per-corpus and per-benchmark expansion of the main results in \autoref{tab:main-compact}.} Per-corpus held-out NLL on four out-of-distribution corpora (WikiText, C4, Pile-Books3, GovReport) and per-benchmark zero-shot accuracy on five standard tasks (ARC-Easy, HellaSwag, PIQA, SciQ, LAMBADA), averaged into Corpus NLL and Task Accuracy in the main text. \textbf{Bold} marks the best per column within each size; \SPS rows are shaded.}
    \label{tab:main-full}
\end{table}

\section{Seed Variance and Statistical Tests}\label{app:seeds}

\begin{figure}[ht]
    \centering
    \includegraphics[width=0.55\textwidth]{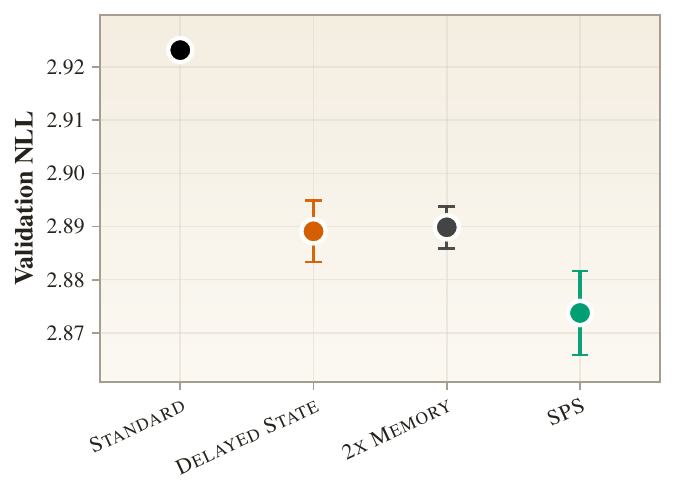}
    \caption{\textbf{Seed-level robustness at S, 10B.} Final FineWeb-Edu validation NLL across $n{=}3$ seeds per method (the headline run plus seed $0$ and seed $1$). Bars are $95\%$ confidence intervals (Student-$t$, $t_{0.025,2}\approx 4.30$).}
    \label{fig:seed_variance}
\end{figure}

Re-training each variant at every scale across multiple seeds is computationally prohibitive at pretraining cost, so we run a focused seed-robustness check at the S, $10$B setting. Each of the four variants (\Standard, \DelayedState, \Twomem, \SPS) is re-trained with three seeds: the headline run plus seed $0$ and seed $1$. The seeds vary both the training-data ordering (the order in which packed sequences are streamed by the data loader) and the weight-initialization random seed; all other hyperparameters are held fixed at their main-table values. \autoref{fig:seed_variance} shows the mean and $95\%$ confidence interval of the final FineWeb-Edu validation NLL across the three seeds; the confidence interval is computed from the Student-$t$ distribution with the multiplier $t_{0.025,\,2}\approx 4.30$ for $n=3$.

We test whether \SPS\ improves over each baseline by a one-sided Welch's $t$-test against the alternative ``\SPS\ has lower validation NLL''. All three baselines reject the null at $p<0.005$ even with the small-$n$ Student-$t$ penalty:

\begin{center}
\begin{tabular}{lccc}
\toprule
Comparison & gap (NLL) & $t$ & $p$ \\
\midrule
\SPS\ vs.\ \Standard      & $-0.0495$ & $-26.6$ & $4.4{\times}10^{-4}$ \\
\SPS\ vs.\ \DelayedState  & $-0.0153$ & $-6.78$ & $1.7{\times}10^{-3}$ \\
\SPS\ vs.\ \Twomem        & $-0.0161$ & $-7.88$ & $2.2{\times}10^{-3}$ \\
\bottomrule
\end{tabular}
\end{center}

\autoref{tab:main-compact} also suggests that \DelayedState\ and \Twomem\ end up at indistinguishable validation NLL despite very different memory footprints. We confirm this with a two one-sided test (TOST) for equivalence within $\pm 0.01$ NLL: TOST $p=3.4{\times}10^{-3}$, so the two are statistically equivalent at this scale within a margin well below the gap to either \SPS\ ($0.0153$) or \Standard\ ($0.034$).

\section{FLOP Accounting}\label{app:flops}

We count FLOPs term by term, following the convention of \citet{hoffmann2022an} and applying the same accounting to every architecture, so that reported compute does not depend on our kernel implementation.
One multiply or one add is one FLOP, so a matmul producing $M$ outputs over a contraction of length $K$ costs $2MK$.
We report forward FLOPs per real input token, and take training FLOPs as $3\times$ forward.

We use the notation of \autoref{app:transformer}: $L$ layers, hidden size $d$, $H$ heads of dimension $d_h = d/H$, feed-forward intermediate size $3d$, context length $T$, and vocabulary $V$.
The per-token forward cost of \Standard\ falls into three groups.
The parameter matmuls contribute
\begin{equation}\label{eq:flops-params}
    \underbrace{2Ld(3d)}_{\text{QKV}} \;+\; \underbrace{2Ld^{2}}_{\text{output projection}} \;+\; \underbrace{2L\cdot 3\cdot d(3d)}_{\text{SwiGLU}} \;=\; 26Ld^{2} \;=\; 2N\;\;,
\end{equation}
where $N = 13Ld^{2}$ is the non-embedding parameter count.
The attention score contributes
\begin{equation}\label{eq:flops-attn}
    \underbrace{2LTd}_{\text{QK logits}} \;+\; \underbrace{3LHT}_{\text{softmax}} \;+\; \underbrace{2LTd}_{\text{value reduction}}\;\;.
\end{equation}
The weight-tied unembedding of \autoref{eq:unembed} contributes $2d\lvert V\rvert$, with $\lvert V\rvert = 50{,}257$.
The input embedding is a row gather, and we assign it no FLOPs.

\SPS, \DelayedState, and \Twomem\ run one forward pass over the interleaved length-$2T$ sequence of \autoref{eq:interleaved}, which changes two of the three groups.
Every parameter matmul in \autoref{eq:flops-params} runs twice per real token, giving a multiplier of $2$.
The attention score scales with the number of query-key interactions, and the sliding window of \autoref{eq:sps-mask} is defined on the key slot: both streams attend to all $T$ persistent entries and to the $w$ recent \texttt{<predict>} entries, so the $2T$ queries induce a multiplier of
\begin{equation}\label{eq:flops-score-mult}
    \frac{2T(T+w)}{T^{2}} \;=\; 2 + \frac{2w}{T}\;\;,
\end{equation}
which is $2.031$ at $w{=}64$ and $T{=}4096$, and exactly $4$ for \Twomem, whose window spans the full context.
The unembedding is charged once for every architecture, since only real tokens emit a prediction.

\autoref{tab:flops} reports the resulting costs.
The \SPS\ ratio grows with scale, from $1.72\times$ at XS to $1.97\times$ at XL, as the parameter matmuls come to dominate the attention and unembedding terms.
\DelayedState\ costs exactly what \SPS\ costs: the two differ only in which slot carries the persistent state, which leaves every term above unchanged.

\begin{table}[ht]
    \centering
    \small
    \setlength{\tabcolsep}{6pt}
    \renewcommand{\arraystretch}{1.2}
    \begin{tabular}{@{}l S[table-format=1.4e1] S[table-format=1.4e1] S[table-format=1.3] S[table-format=1.3] S[table-format=1.3]@{}}
        \toprule
        & {\Standard\ forward} & {\Standard\ training} & \multicolumn{3}{c}{Forward ratio vs.\ \Standard} \\
        \cmidrule(lr){2-3} \cmidrule(lr){4-6}
        Size & {(FLOPs/token)} & {(FLOPs/token)} & {\SPS} & {\DelayedState} & {\Twomem} \\
        \midrule
        XS & 1.7388e8  & 5.2165e8  & 1.716 & 1.716 & 2.485 \\
        S  & 4.1398e8  & 1.2420e9  & 1.825 & 1.825 & 2.552 \\
        M  & 1.1646e9  & 3.4938e9  & 1.923 & 1.923 & 2.611 \\
        L  & 2.4260e9  & 7.2781e9  & 1.957 & 1.957 & 2.577 \\
        XL & 4.6287e9  & 1.3886e10 & 1.974 & 1.974 & 2.515 \\
        \bottomrule
    \end{tabular}
    \vspace*{1em}
    \caption{\textbf{Per-token FLOP costs and per-architecture ratios.} Forward FLOPs per real input token for \Standard, the corresponding training cost at $3\times$ forward, and the ratio each variant pays at the same scale.}
    \label{tab:flops}
\end{table}

\begin{figure}[ht]
    \centering
    \includegraphics[width=\textwidth]{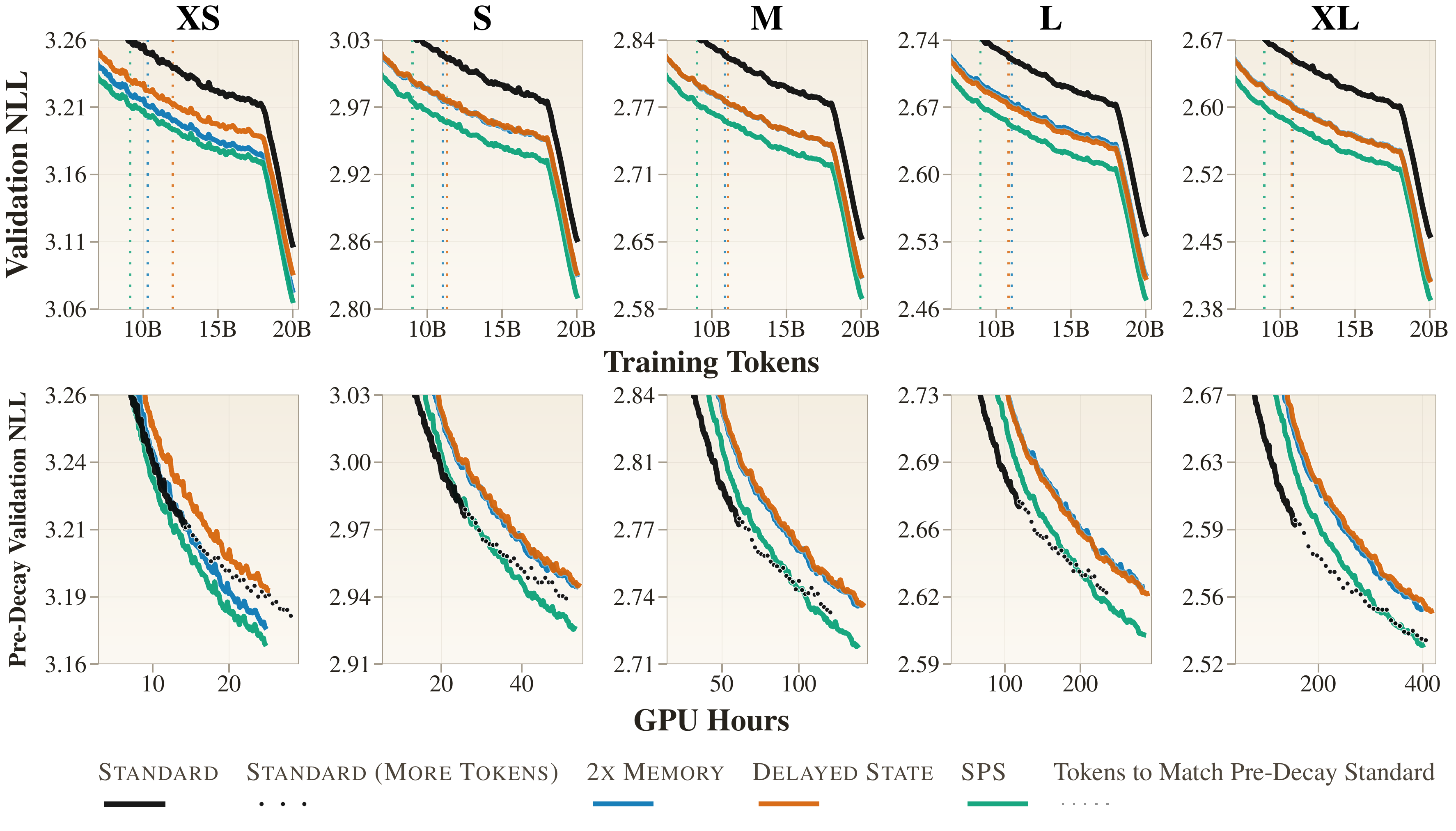}
    \caption{\textbf{Measured GPU-hours view of \autoref{fig:history_combined}.} FineWeb-Edu validation NLL vs.\ tokens seen (top) and measured GPU-hours (bottom). The top row includes LR cool down, hence the drop at the end. Unlike the training-FLOPs axis, this view reflects kernel efficiency and hardware, and is specific to the implementation and GPUs described in \autoref{sec:experimental-setup}.}
    \label{fig:history_combined_gpu}
\end{figure}

Measured training time puts \SPS\ at higher cost than FLOPs alone would predict.
This overhead comes from our attention kernels, and a better implementation would narrow it.
We therefore report the analytic ratio wherever FLOPs appear on an axis, and report measured GPU-hours separately in \autoref{fig:history_combined_gpu}.

\section{Compute-Optimal Comparison}\label{app:compute-optimal}

\begin{figure}[ht]
    \centering
    \includegraphics[width=0.7\textwidth]{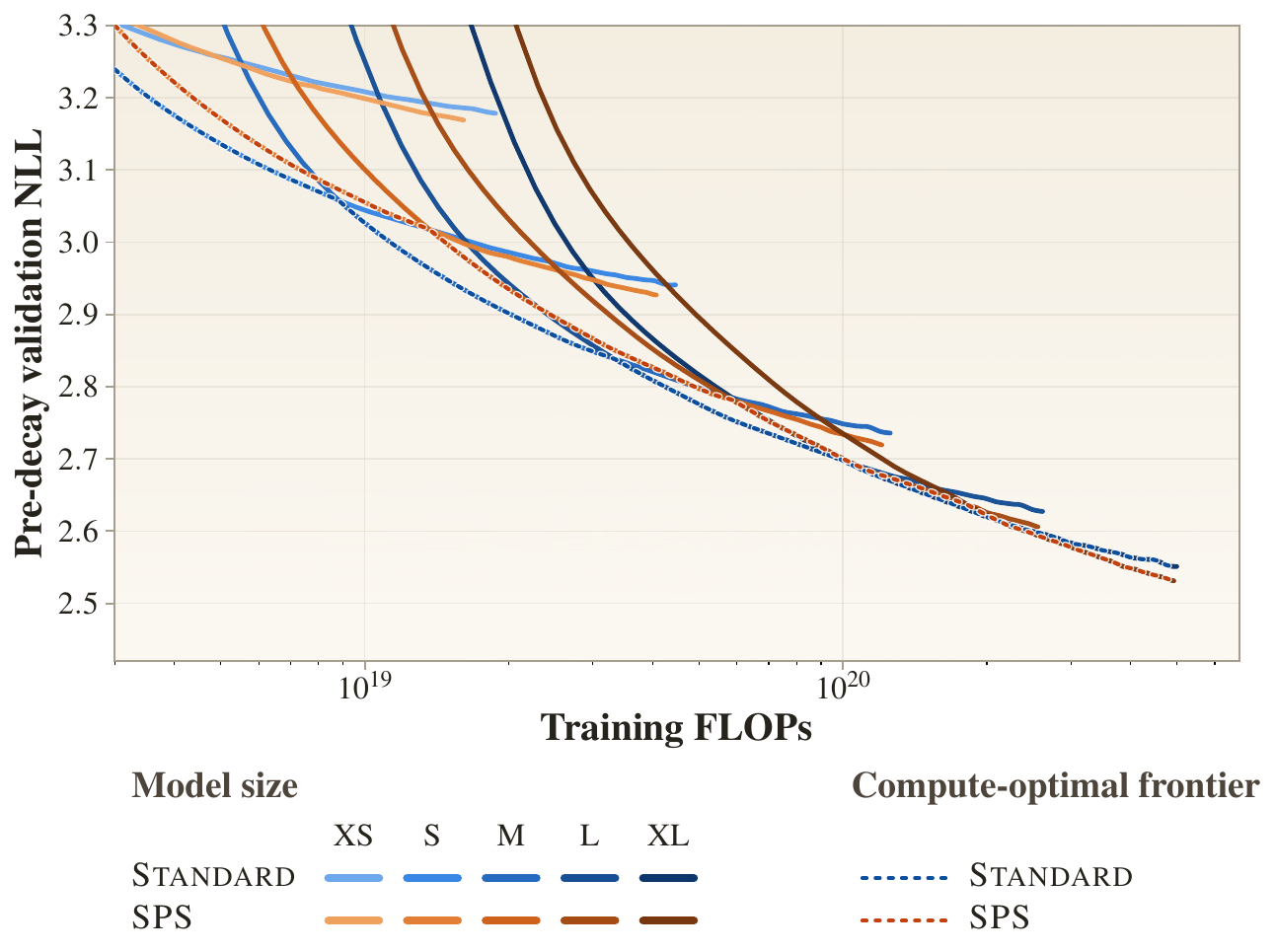}
    \caption{\textbf{Pre-decay validation NLL vs.\ training FLOPs.} Every \Standard\ and \SPS\ run, coloured by model size. The dotted lines are each family's compute-optimal frontier, the lower envelope over model sizes at each budget.}
    \label{fig:loss_vs_compute}
\end{figure}

A fixed compute budget can be spent on a larger model or on more tokens, so we also compare the two families at matched training FLOPs.
\autoref{fig:loss_vs_compute} plots pre-decay validation NLL against training FLOPs for every run, and we read each family's compute-optimal frontier from it as the best NLL available at each budget.
\Standard\ leads at low budgets, where the compute is better spent on a larger model, and its advantage shrinks monotonically as the budget grows (\autoref{tab:compute-optimal}).
The frontiers cross at $2.1\times10^{20}$ FLOPs, and \SPS\ leads by $0.020$ nats at our largest budget.
At $2.6\times10^{20}$ FLOPs \SPS's XL run overtakes \Standard's XL having seen half as many tokens.
Model families are commonly trained for a single large token budget across sizes, well past the compute-optimal point, so the token-matched comparison of \autoref{fig:history_combined} remains the more practically relevant one.

\begin{table}[ht]
    \centering
    \small
    \setlength{\tabcolsep}{6pt}
    \renewcommand{\arraystretch}{1.2}
    \begin{tabular}{@{}l S[table-format=1.1e1] S[table-format=1.3] S[table-format=1.3] S[table-format=+1.3]@{}}
        \toprule
        & & \multicolumn{2}{c}{Best NLL at this budget ($\downarrow$)} & \\
        \cmidrule(lr){3-4}
        Budget & {Training FLOPs} & {\Standard} & {\SPS} & {$\Delta$ ($\downarrow$)} \\
        \midrule
        XS final & 1.6e19 & {2.934\,\scriptsize {(M, 4.6B)}}  & {2.979\,\scriptsize {(M, 2.4B)}}  & +0.045 \\
        S final  & 4.1e19 & {2.806\,\scriptsize {(L, 5.6B)}}  & {2.825\,\scriptsize {(M, 6.1B)}}  & +0.019 \\
        M final  & 1.2e20 & {2.674\,\scriptsize {(XL, 8.7B)}} & {2.677\,\scriptsize {(L, 8.5B)}}  & +0.003 \\
        \rowcolor{winrow}
        L final  & 2.6e20 & {2.598\,\scriptsize {(XL, 18.4B)}} & {2.593\,\scriptsize {(XL, 9.3B)}} & -0.005 \\
        \bottomrule
    \end{tabular}
    \vspace*{1em}
    \caption{\textbf{Compute-optimal comparison at matched training FLOPs.} At each budget we report the best pre-decay validation NLL available to each family, with the model size that achieves it and the tokens that model has seen. Budgets are the last budget at which each size is still training, so two sizes of each family are available at every row. $\Delta$ is \SPS\ minus \Standard, and negative favours \SPS.}
    \label{tab:compute-optimal}
\end{table}

\section{Attention Patterns of the Variants}\label{app:attention}

\begin{figure}[ht]
    \centering
    \includegraphics[width=\textwidth]{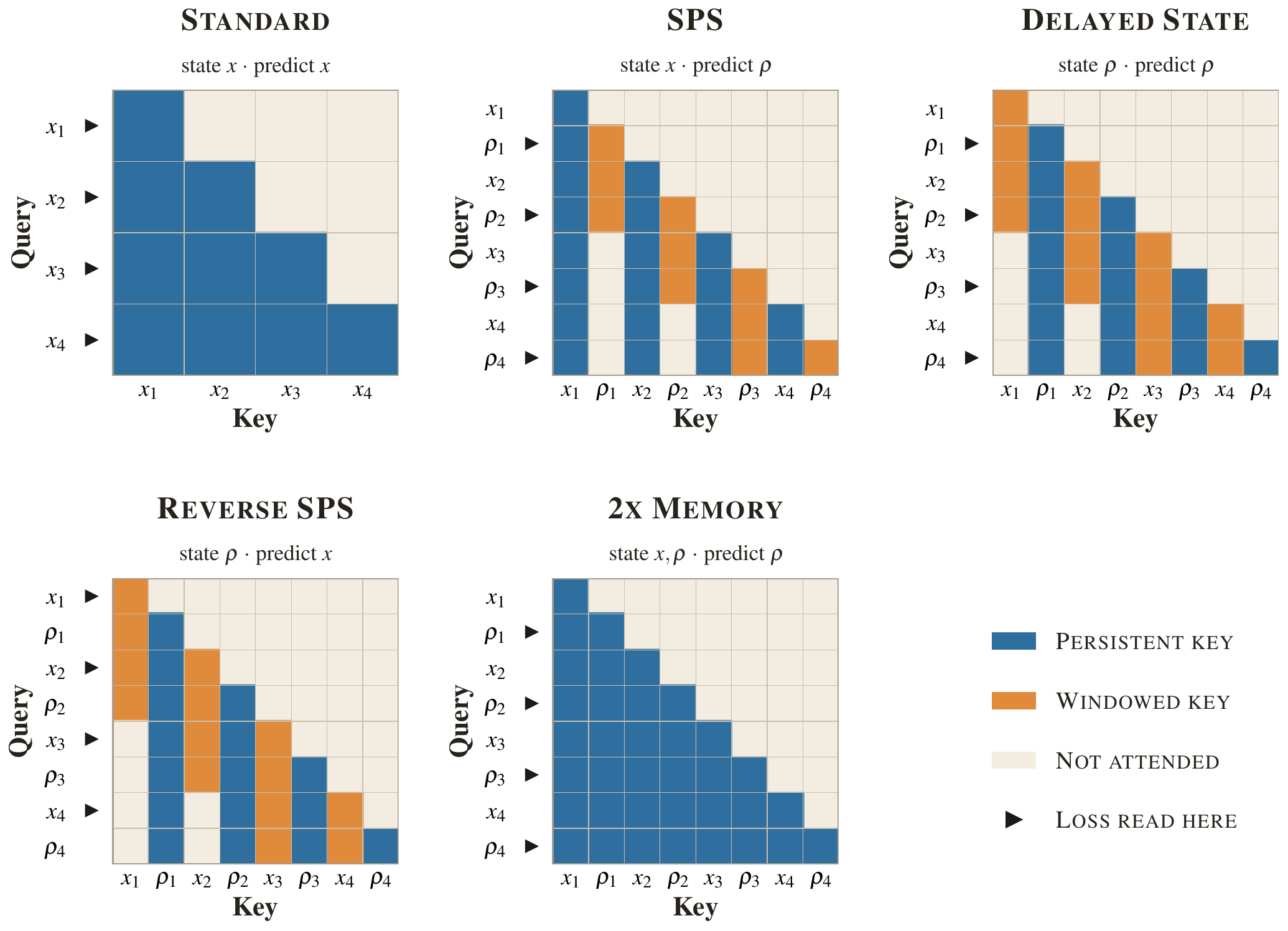}
    \caption{\textbf{Attention mask and prediction slot for each variant.} Rows are queries and columns are keys, drawn with $T{=}4$ tokens and window $w{=}1$. Persistent keys are never evicted; windowed keys are visible only while they remain within $w$. A marker on the left of a row indicates that the next-token loss is read from that slot. \DelayedState\ and \Inverse\ share the same mask and differ only in that marker.}
    \label{fig:attention_masks}
\end{figure}

Every predict-token variant interleaves a \texttt{<predict>} token after each input token as in \autoref{eq:interleaved}, and computes the loss only at its designated prediction slot.
The variants differ in two choices: which slot holds the persistent keys, and which slot the loss is read from.
\SPS\ keeps the input slot persistent and reads the loss from the \texttt{<predict>} slot, so the two roles fall on different streams.
\DelayedState\ keeps the \texttt{<predict>} slot persistent and reads the loss from that same slot, so one stream carries both roles.
\Inverse\ keeps the \texttt{<predict>} slot persistent as well, but reads the loss from the input slot.
\DelayedState\ and \Inverse\ therefore share an identical attention mask and differ only in the slot the prediction is made at.
\Twomem\ keeps both slots persistent, which is \SPS\ with the window widened to the full context.
\autoref{fig:attention_masks} draws all five.

\end{document}